\documentclass[conference]{IEEEtran}
\IEEEoverridecommandlockouts
\usepackage{cite}
\usepackage{amsmath,amssymb,amsfonts}
\usepackage{algorithmic}
\usepackage{caption}
\usepackage{graphicx}
\usepackage{textcom	p}
\usepackage{xcolor}
\usepackage{graphicx,subfig,epstopdf,booktabs,amsmath,paralist,float,lineno,caption}
\usepackage{xfrac}
\def\BibTeX{{\rm B\kern-.05em{\sc i\kern-.025em b}\kern-.08em
    T\kern-.1667em\lower.7ex\hbox{E}\kern-.125emX}}
\begin{document}

\title{Multi-scale Cross-form Pyramid Network \\ for Stereo Matching
\thanks{*Mingyi He, corresponding author, Email: myhe@nwpu.edu.cn}
}

\author{\IEEEauthorblockN{Zhidong Zhu, Mingyi He*, Yuchao Dai, Zhibo Rao, Bo Li}
\IEEEauthorblockA{\textit{School of Electronics and Information} \\
\textit{Northwestern Polytechnical University}\\
Xi'an, China \\
Email: myhe@nwpu.edu.cn}
}

\maketitle

\begin{abstract}
Stereo matching plays an indispensable part in autonomous driving, robotics and 3D scene reconstruction. We propose a novel deep learning architecture, which called CFP-Net, a cross-form pyramid stereo matching network for regressing disparity from a rectified pair of stereo images. The network consists of three modules: Multi-Scale 2D local feature extraction module, Cross-form spatial pyramid module and Multi-Scale 3D Feature Matching and Fusion module. The Multi-Scale 2D local feature extraction module can extract enough multi-scale feature. The Cross-form spatial pyramid module aggregates the context information in different scales and locations to form a cost volume. Moreover, it is been proved to be more effective than SPP and ASPP in ill-posed regions. The Multi-Scale 3D feature matching and fusion module is prove to regularize cost volume using two parallel 3D deconvolution structure with two different receptive fields. Our proposed method has been evaluated on the Scene Flow and KITTI datasets. It achieves state-of-the-art performance on the KITTI 2012 and 2015 benchmarks.
\end{abstract}


\section{Introduction}
Stereo matching, as a classic problem of 3D scene reconstruction, has been extensively studied in recent years. Binocular stereo matching is mainly used in robot vision, non-contact measurement, pose detection and control of mechanical systems and virtual reality\cite{LI2018328}\cite{Li2015Depth}. The binocular camera captures the left and right viewpoint images of the same scene and uses the stereo matching algorithm to obtain the disparity map and corresponding depth map. Given a pair of rectified stereo images, the purpose of the stereo matching algorithm is to find the geometric relationship of the same pixels on the left and corresponding right image.

Traditional stereo methods, such as belief propagation\cite{Klaus2006Segment}, semi-global matching\cite{4359315} and max-flow\cite{Kolmogorov2001Computing}, are well studied following stereo pipeline with limited performance. Traditional stereo methods, such as belief propagation\cite{Klaus2006Segment}, semi-global matching\cite{4359315} and max-flow\cite{Kolmogorov2001Computing}, are well studied following stereo pipeline with limited performance. Scharsteim and Sceliski(2002) proposed that a typical stereo algorithm consists of four steps: matching cost computation, cost aggregation, optimization, and disparity refinement. The point of this work is on computing a better matching cost.

\begin{figure*}[htbp]
   \centering
   \includegraphics[width=1.0\textwidth]{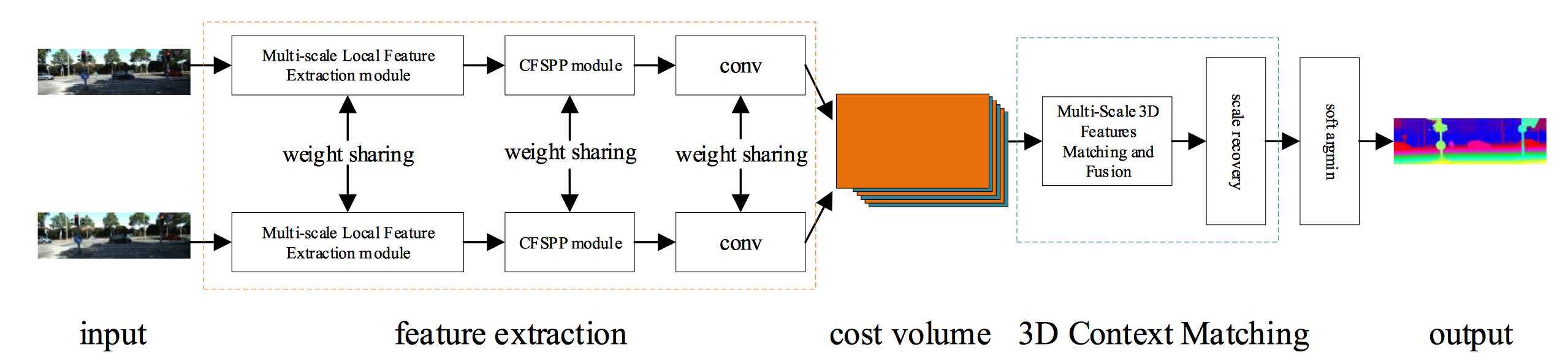} 
   \caption{Architecture of our CFP-Net}
   \label{fig:Networks}
\end{figure*}

With the fast development of deep learning methods on semantic understanding, more and more stereo matching methods uses convolution neural networks(CNNs).
The key and difficult point is the utilization of the integral global context information. Recent studies\cite{Chang2018Pyramid}\cite{Pang2017Cascade} attempt to solve this problem by combining semantic information to refine matching cost volumes. CRLNet\cite{Pang2017Cascade} utilizes two stages of convolutional neural networks with hour-glass structure and embeds the residual learning mechanism across multiple scales. PSM-Net\cite{Chang2018Pyramid} uses Spatial pyramid pooling(SPP) and dilated convolution to exploit global context information effectively with enlarged receptive fields. However, deep learning methods are still not effective in dealing with the problem of disparity estimation for inherently ill-posed regions. In this work, inspired by the spatial pyramid pooling(SPP) in PSM-Net\cite{Chang2018Pyramid} and the dilated convolution structure in DeepLab v3+\cite{Chen2018Encoder}, we redesign a cross-form spatial pyramid pooling architecture(CFSPP) to enlarge the receptive fields from multiple perspectives in order to extends pixel-level features to region-level features with different scales of receptive fields.

In this work, we propose a novel parallel cross-form pyramid network(CFP-Net), which can exploit global context information effectively and completely in stereo matching. We extend the spatial pyramid module to cross-form level with four parallel pyramid pooling blocks, containing parallel dilated convolutions layer and average pooling layer. Moreover, we exploit global context information effectively from the perspective of the whole image by our encoder-decoder architecture described by following sections. We conducted many comparative experiments, and the results of the experiments prove that CFP-Net has a good effect on stereo matching.
 
The contributions of this paper can be summarized as follows: 

1) We propose an end-to-end stereo matching learning framework without any post-processing.

2) We design a cross-form context spatial pyramid architecture(CFSP) to incorporate global information and local information into image features with enlarged receptive fields. 

3) We redesign Multi-Scale 3D features matching and fusion convolutional neural networks module to increase the utilization of context information in multi-scale cost volume. 

4) We achieve the state-of-art performance in the KITTI 2012 and 2015 benchmarks.

\section{Proposed method}
In this section, we will give a detailed description about our proposed Parallel multi-scale cross-form pyramid network(CFP-Net), which contains three parts: multi-scale local feature extraction module, cross-form spatial pyramid pooling module and multi-scale 3D features matching and fusion module. The architecture of our CFP-Net is presented in Fig. \ref{fig:Networks}.

\subsection{Multi-Scale Local Feature Extraction}
\begin{figure*}[htbp]
   \centering
   \includegraphics[width=0.9\textwidth]{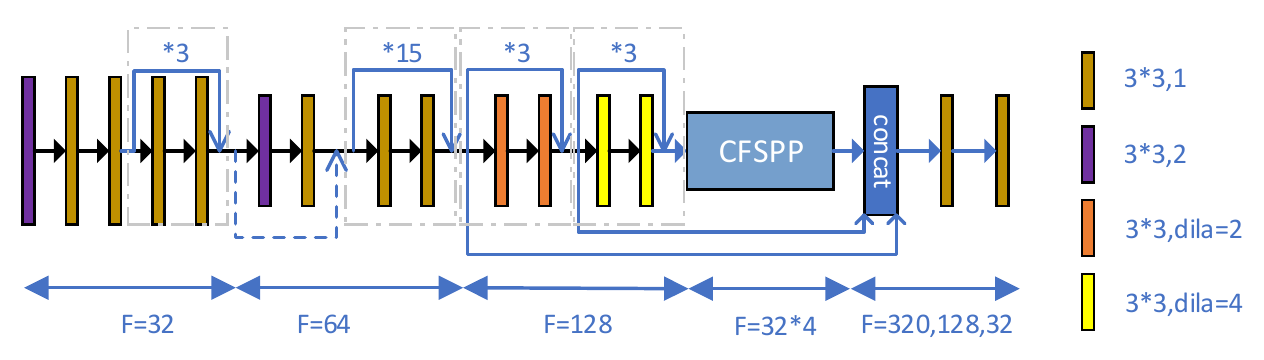} 
   \caption{Architecture of our Feature Extraction Module}
   \label{fig:Architecture of our Feature Extraction Module}
\end{figure*}

It is diffusely recognized that feature descriptors can better capture local context, thus more robust to photometric differences(occlusion, non-lambertian lighting effects and perspective effects)\cite{Zhong2017}. The multi-scale feature extraction module extracts multi-scale features from the stereo image pair. It contains a convolution layer with channel of 32 and kernel size of 3 to improve the dimension of given inputs, and multi-scale serial networks with different scale of 32, 64 and 128. The specific feature extraction module is shown in Fig.\ref{fig:Architecture of our Feature Extraction Module}.

Specifically, we design our local feature extraction module through a series of interconnected 2D convolutional operators. It total possesses the 48 convolution layers consisted of 14 basic blocks with different channels. The main architecture of the local feature extraction module can be divided into following parts:

1) A common 2D convolutional layer \emph{conv0} with channel of 32 and kernel size of 3. The outputs of \emph{conv0} are regarded as the inputs of networks after \emph{conv0}.

2) There are two convolutional layers with stride of 1 in the basic block. And there are 3 basic blocks with channel of 32, 15 basic blocks with channel of 64 and 3 basic blocks with channel of 128 serially. Especially, the first convolutional layer of the first basic block of each part has a stride of 2.

\subsection{Cross-Form Context Pyramid}
Obviously, we could not gather enough context information just by local feature extraction module. It is proved by many related works that image features rich with object context information, such as ill-posed or textureless regions and objects with arbitrary sizes, can benefit correspondence estimation. In addition to this, a multi-scale context representation is sufficient and essential for stereo matching. Hence context pyramid architecture has been referred in many recent stereo matching works since it was proposed by\cite{He2015}. It can offer fine-grained details to make the generated disparity map richer and combine context cues with global and local priors.

\begin{figure}[!htb]
   \centering
   \includegraphics[width=0.45\textwidth]{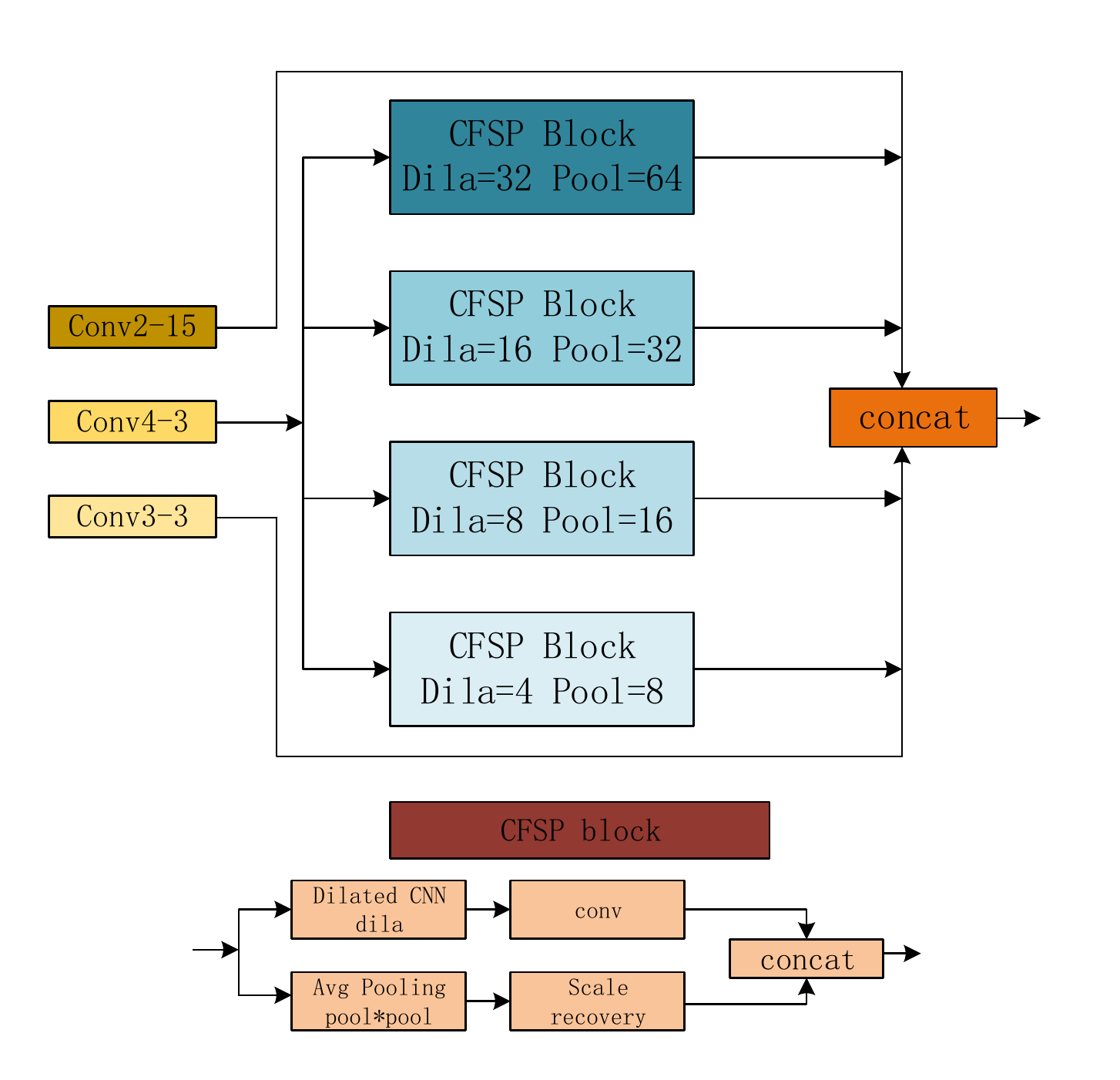} 
   \caption{The basic architectures of CFSPP.}
   \label{fig:CFSPP}
\end{figure}

\begin{figure*}[htbp]
   \centering
   \includegraphics[width=0.9\textwidth]{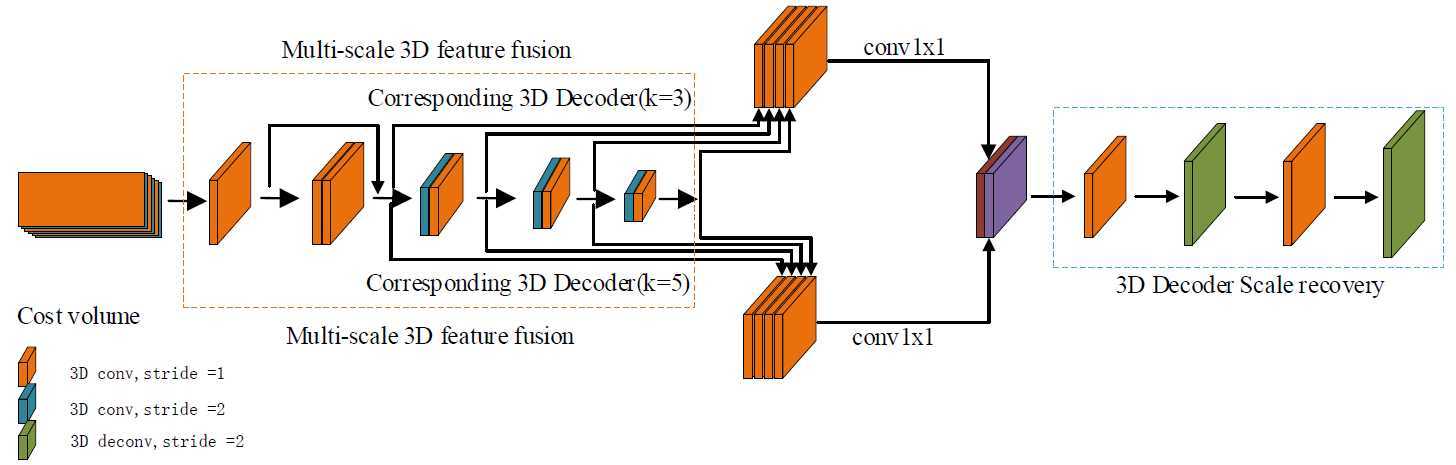} 
   \caption{Architecture of our 3D CNN}
   \label{fig:3DCNN}
\end{figure*}

\begin{figure*}[!htb]
	\centering
	\includegraphics[width=0.31\linewidth]{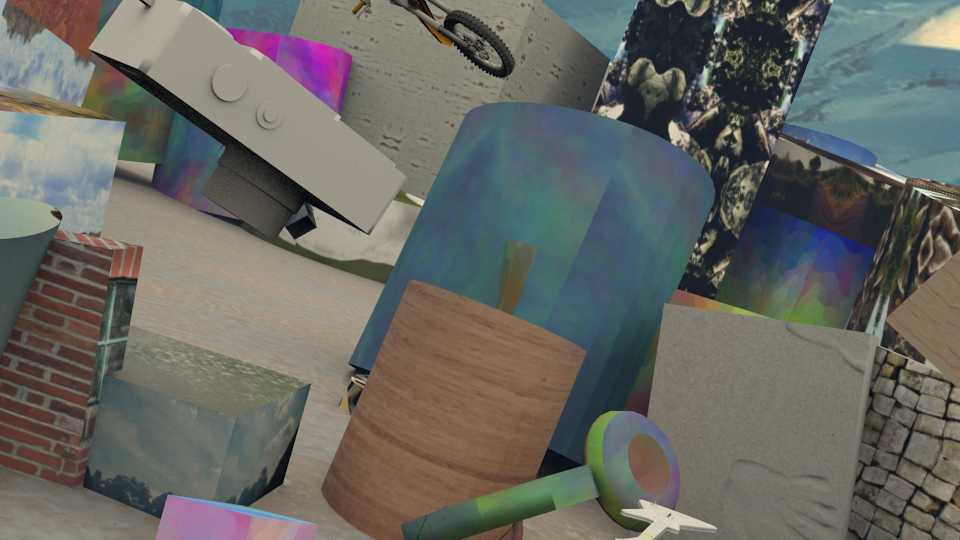}
	\includegraphics[width=0.31\linewidth]{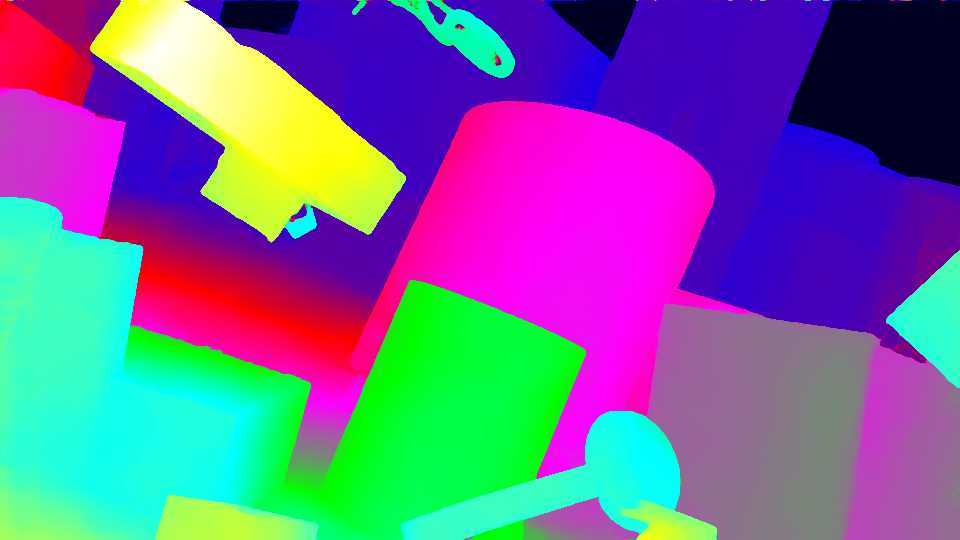}
	\includegraphics[width=0.31\linewidth]{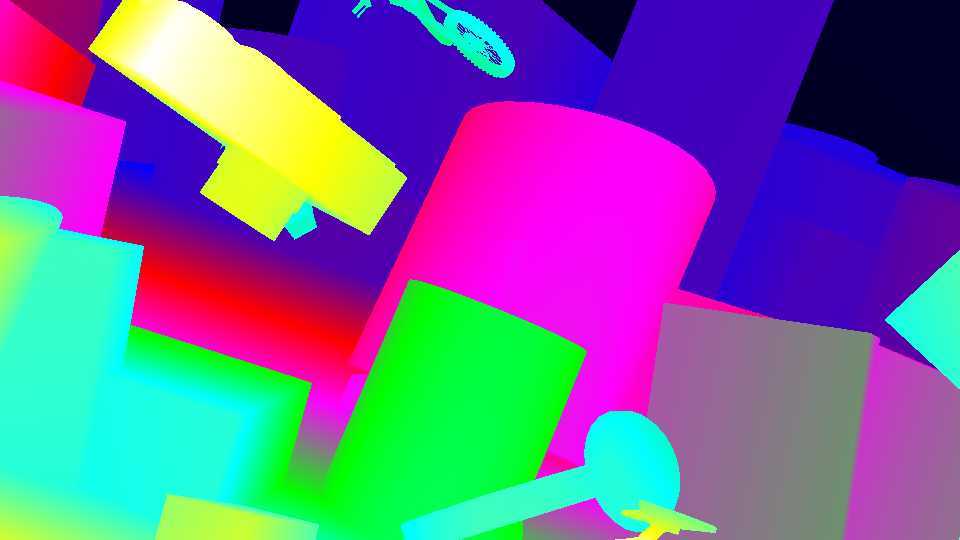}
	\includegraphics[width=0.31\linewidth]{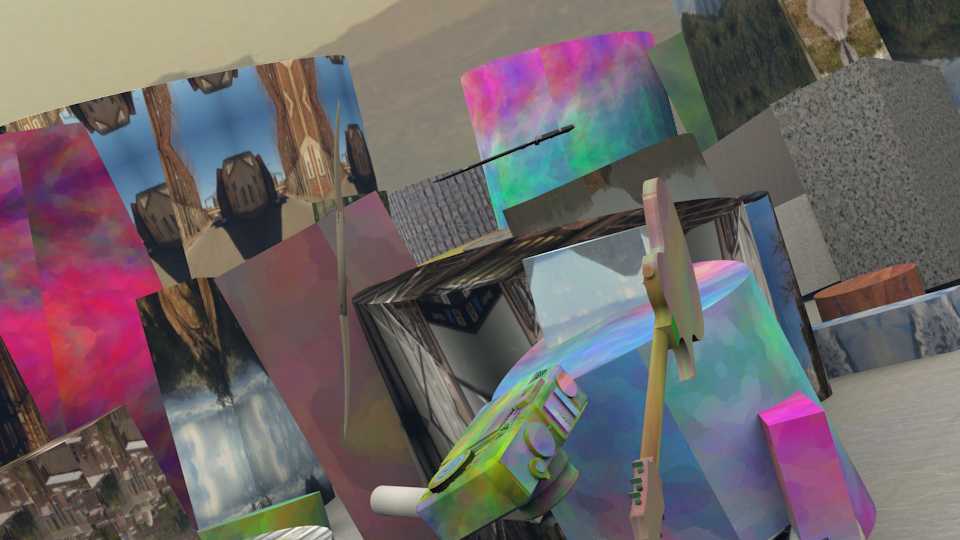}
	\includegraphics[width=0.31\linewidth]{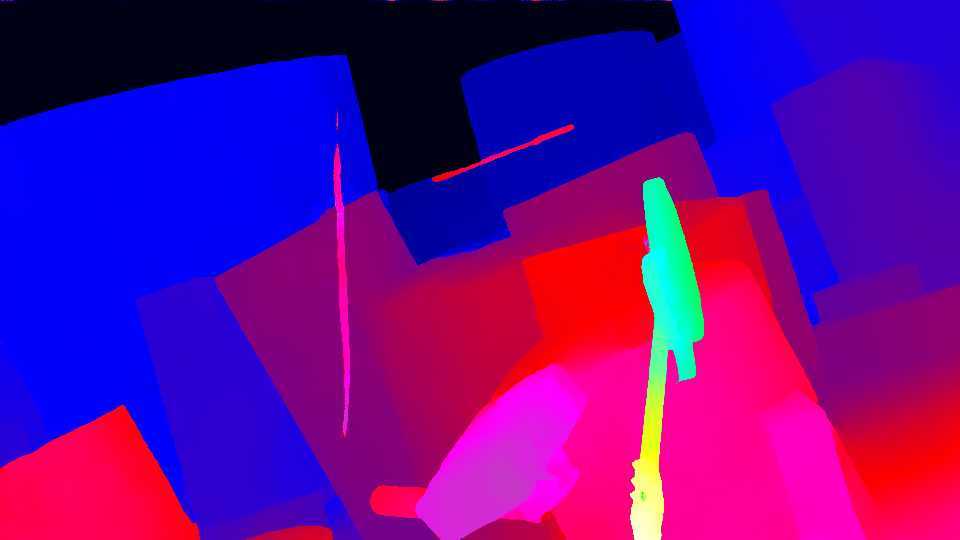}
	\includegraphics[width=0.31\linewidth]{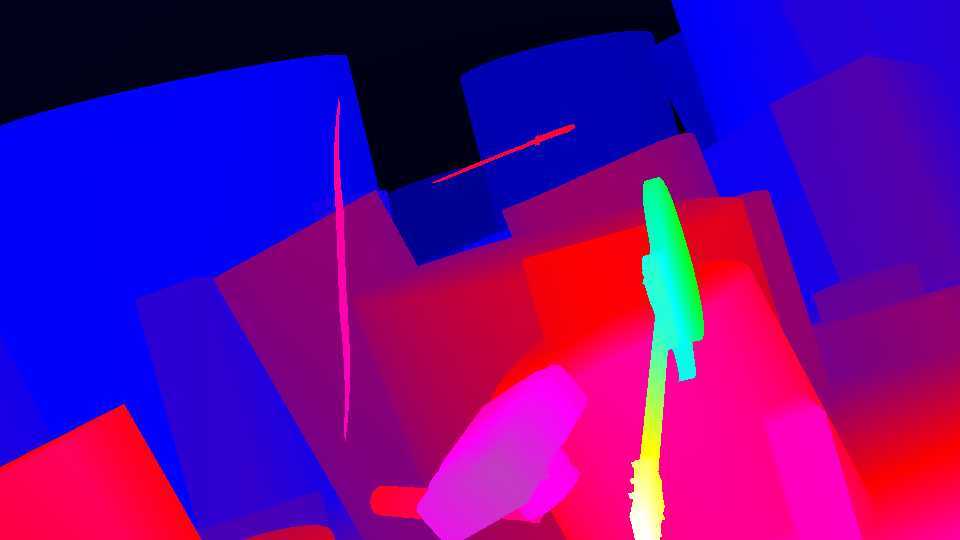}
	\caption{\textbf{SceneFlow test data qualitative results.} From left: left stereo input image, our disparity map, our groundtruth map}
	\label{fig:SceneFlow}
\end{figure*} 

\textbf{\small Proposed Cross-form Context Pyramid.} In this work, as shown in Fig. \ref{fig:CFSPP}. our context pyramid is formulated as a module of four parallel cross-form branches corresponding to four pyramid levels. The context scale is decreasing from top to bottom of the context pyramid. Each cross-form branch consists of two small parallel branches named dilated branch and pooling branch respectively, following a convolution layer for contacting two outputs.

(i) Dilated branch. The first layer is a \(3 \times 3\) dilated conv layer and the second layer is an \(1 \times 1\) conv layer.

(ii) Pooling branch. The first layer is an average pooling layer with corresponding pyramid scale. And the second layer is the same as the dilated branch, following the bilinear interpolation in order to maintain the size of output same as the local stereo volume.

Inspired by \cite{Chang2018Pyramid}, we design four fixed-size average pooling pyramid blocks for each pooling branch: \(64 \times 64\), \(32 \times 32\), \(16 \times 16\), and \(8 \times 8\). Correspondingly, we also design four fixed-size dilation for each dilated branch in order: 32, 12, 8, 4. In \cite{Chen2018Encoder}, Atrous convolution spatial context pyramid is introduced to adjust filter's field-of-view in order to capture multi-scale information and generalize standard convolution operation.

For comparison, we set another two contrast context pyramid structures in \cite{Chang2018Pyramid} and \cite{Chen2018Encoder}. Atrous Spatial Pyramid Pooling(ASPP)\cite{Chen2018Encoder} remove all pooling branches of the cross-form context pyramid with the remain parameter abiding. Spatial Pyramid Pooling(SPP)\cite{Chang2018Pyramid} remove all dilated branches of the cross-form context pyramid with the remaining parameter abiding. We conduct multiple comparative experiments to compare these three context pyramids in section 4.2.

\subsection{Feature Volume}
Traditional stereo matching methods just use a simple way to concatenate the stereo result. And recent GC-Net\cite{Kendall2017} methods concatenate the left and right features to form a cost volume. Following \cite{Chang2018Pyramid}, we concatenate left feature maps with their corresponding right feature maps across each disparity level with our CFSPP to form a cost volume with size \(height \times width\ \times disparity\ \times size\). 

\subsection{Multi-scale 3D Features Matching and Fusion Module}
Our CFSPP module makes a great contribution to stereo matching from different perspectives by parallel dilated branch and pooling branch. We capture both the global context information with the receptive field of the whole image and local context information with the receptive field of different scales. Hence, we redesign the MP-SOD method purposed by Zhang\cite{Zhang2018Multi} as our 3D Features matching and fusion (encoder-decoder) module in order to aggregate the feature information along both the disparity dimension and spatial dimensions. There are mainly two changes comparing with MP-SOD that we 1) transform all 2D convolutional layers to 3D convolutional layers or 3D deconvolutional layers. 2) introduce 3D deconvolutional layers as scale recovery module to make the concatenated feature map upsample to the same dimension as the input image. The spatial resolution of the output feature map is increased three times under our structure.

As shown in Fig. \ref{fig:3DCNN}, the multi-scale 3D features matching and fusion module is divided into two parts. The first part consists of a series of three-dimension convolutional layers with increasing size of the receptive fields. For the multi-scale features matching module, four 3D convolutional layers with different scales are utilized, where the channels are set as 32, 32, 64, 64. Moreover, each feature map is fed through a series of corresponding 3D deconvolutional layers with different kernel sizes of 3 or 5 respectively to control the size of the receptive field. and upsampled to the same dimension as cost volume. The second part, which leads to the fusion of context information extracted through 3D convolution with different scales, consists of two parallel concatenated feature maps, which are supplied to a two-channel feature map by a 3D convolutional layer, following a scale recovery module.

\begin{table*}
\centering
\begin{tabular}{l|ccc|ccc|ccc} \hline
                   & \multicolumn{3}{c|}{$>3$ px} & \multicolumn{3}{c|}{$>4$ px} & \multicolumn{3}{c}{$>5$ px} \\ 
                   & Noc & All & Refl & Noc & All & Refl & Noc & All & Refl    \\ \hline
GC-NET\cite{Kendall2017} 		& 1.77 & 2.30 & 12.80 & 1.36 & 1.77 & 9.77 & 1.12 & 1.46 & 7.99	\\
L-ResMatch\cite{Shaked2017Improved} 		& 2.27 & 3.40 & 19.71 & 1.76 & 2.67 & 16.52 & 1.50 & 2.26 & 14.52	\\ 
SsSMNet\cite{Zhong2017} 		& 2.30 & 3.00 & 16.59 & 1.82 & 2.39 & 13.21 & 1.53 & 2.01 & 11.08	\\ 
SGM-Net\cite{Seki2017SGM} 		& 2.29 & 3.50 & 18.97 & 1.83 & 2.80 & 15.62 & 1.60 & 2.36 & 13.55	\\ 
PBCP\cite{Seki2016Patch} 		& 2.36 & 3.45 & 20.29 & 1.88 & 2.74 & 16.75 & 1.62 & 2.32 & 14.52	\\ 
Displets v2\cite{Guney2015} 		& 2.37 & 3.09 & 10.41 & 1.97 & 2.52 & 8.02 & 1.72 & 2.17 & 6.61	\\ 
PSMNet\cite{Chang2018Pyramid} 		& 1.49 & 1.89 & 10.18 & 1.12 & 1.42 & 7.29 & 0.90 & 1.15 & 5.64	\\ \hline
CFPNet (proposed network)    & \textbf{1.41} & \textbf{1.83} & \textbf{9.58} & \textbf{1.06} & \textbf{1.37} & \textbf{6.90} & \textbf{0.85} & \textbf{1.10} & \textbf{5.33} \\      
\end{tabular}
	\caption{\textbf{KITTI 2012 test set results} \cite{Menze2015CVPR}. This benchmark contains 195 training and 195 test color image pairs.}
\label{tab:Model Performance on 2012}
\end{table*}

\begin{table*}
\centering
\begin{tabular}{l|ccc|c|c} \hline
                   & \multicolumn{3}{c|}{All Pixels} & Param. & Runtime \\
Model Type & $>1$ px & $>3$ px & $>5$ px  & (M) & (s)    \\ \hline
LFE+SPP+3DCNN			& 5.41 & 4.18 & \textbf{3.44} & 4.22 & 0.8 \\
LFE+ASPP+3DCNN 	& 5.5 & 4.15 & 3.49 & 4.36 & 0.8 \\
LFE(replaced)+CFSPP+3DCNN & 6.04 & 4.92  & 4.13 & 4.24 & 0.7 \\
LFE+CFSPP+3DCNN(replaced) 			& 5.9 & 4.74 & 3.64 & 4.03 & 0.6\\ \hline
LFE+CFSPP+3DCNN (proposed network)    & \textbf{5.37} & \textbf{4.03} & 3.48  & 4.68 & 0.95 \\      
\end{tabular}
	\caption{\textbf{Ablation Experiments on Scene Flow datasets}}
\label{tab:Pivotal Architecture Comparison}
\end{table*}

\begin{table*}
\centering
\begin{tabular}{l|ccc|ccc|c} \hline
                   & \multicolumn{3}{c|}{All Pixels} & \multicolumn{3}{c|}{Non-Occluded Pixels} & Runtime \\
                   & D1-bg   & D1-fg   & D1-all  & D1-bg   & D1-fg   & D1-all  & (s)     \\ \hline
GC-Net\cite{Kendall2017}			& 2.21 & 6.16 & 2.87 & 2.32 & 3.12 & 2.45 & 0.9 \\
SsSMnet\cite{Zhong2017} 	& 2.70 & 6.92 & 3.40 & 2.46 & 6.13 & 3.06 & 0.8 \\
L-ResMatch\cite{Shaked2017Improved} & 2.72 & 6.95 & 3.42 & 2.35 & 5.74  & 2.91 & 48 \\
PBCP \cite{Seki2016Patch} 			& 2.58 & 8.74  & 3.61 & 2.27 & 7.71  & 3.17 & 68  \\
PSMNet\cite{Chang2018Pyramid} & \textbf{1.86} & 4.62 & 2.32 & \textbf{1.71} & 4.31 & 2.14 & \textbf{0.41} \\ 
Displets v2 \cite{Guney2015}& 3.00 & 5.56  & 3.43 & 2.73 & 4.95  & 3.09 & 265 \\ \hline
CFP-Net (our work)					& 1.90 & \textbf{4.39}  & \textbf{2.31} & 1.73 & \textbf{3.92}  & \textbf{2.09} & 0.95 \\      
\end{tabular}
	\caption{\textbf{KITTI 2015 test set results} \cite{Menze2015CVPR}.}
\label{tab:Model Performance on 2015}
\end{table*}

\begin{figure*}[!htb]
	\centering
	\includegraphics[width=0.31\linewidth]{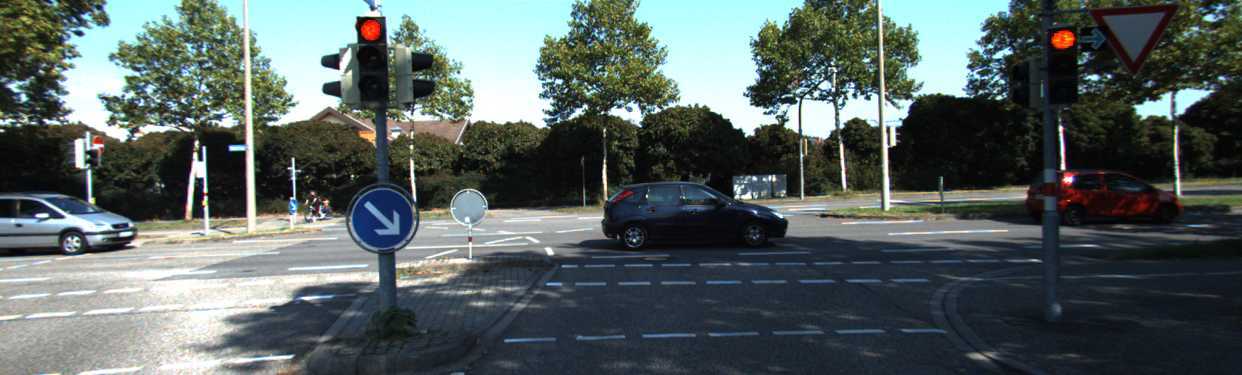}
	\includegraphics[width=0.31\linewidth]{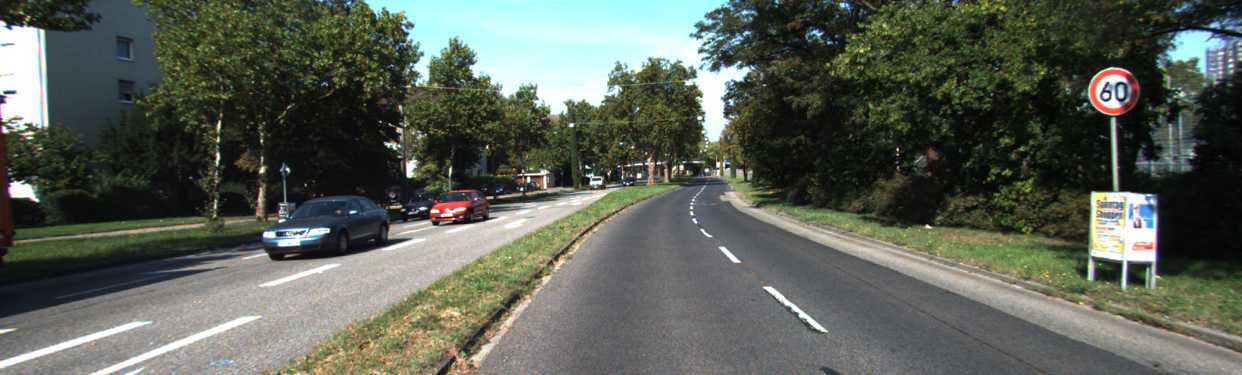}
	\includegraphics[width=0.31\linewidth]{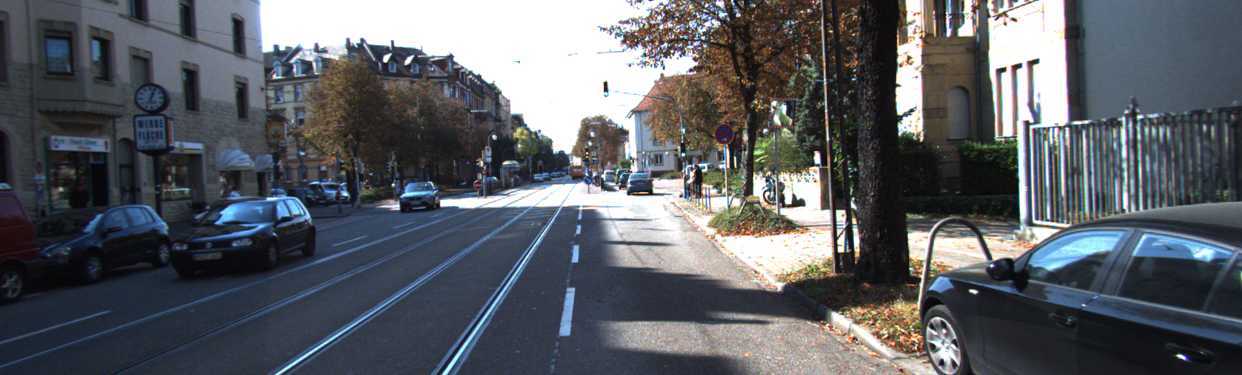}
	\includegraphics[width=0.31\linewidth]{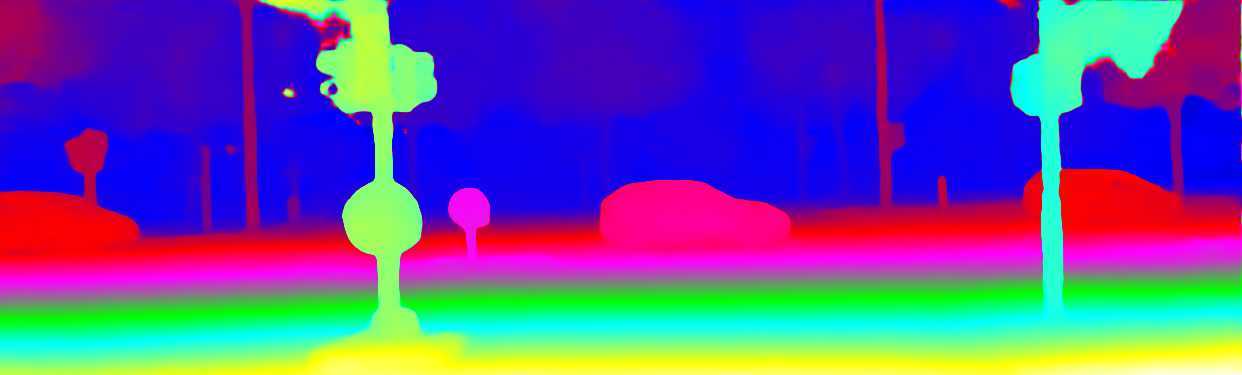}
	\includegraphics[width=0.31\linewidth]{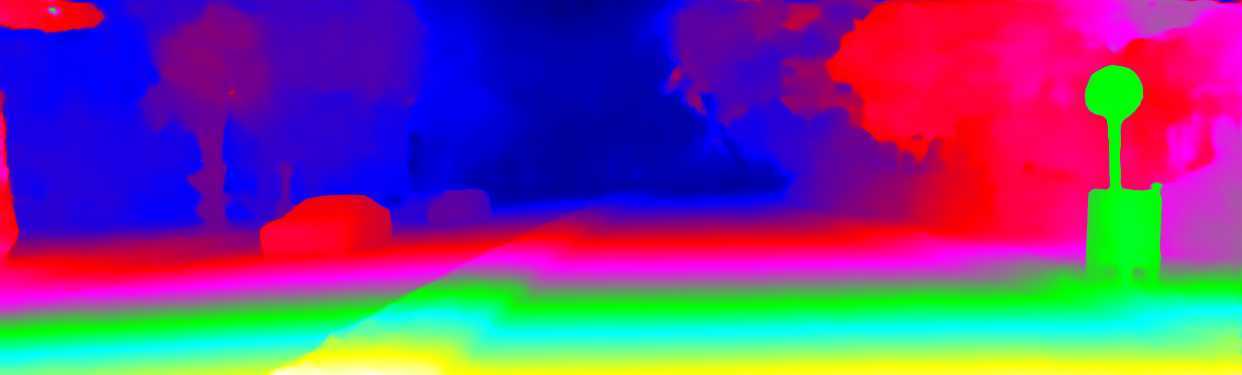}
	\includegraphics[width=0.31\linewidth]{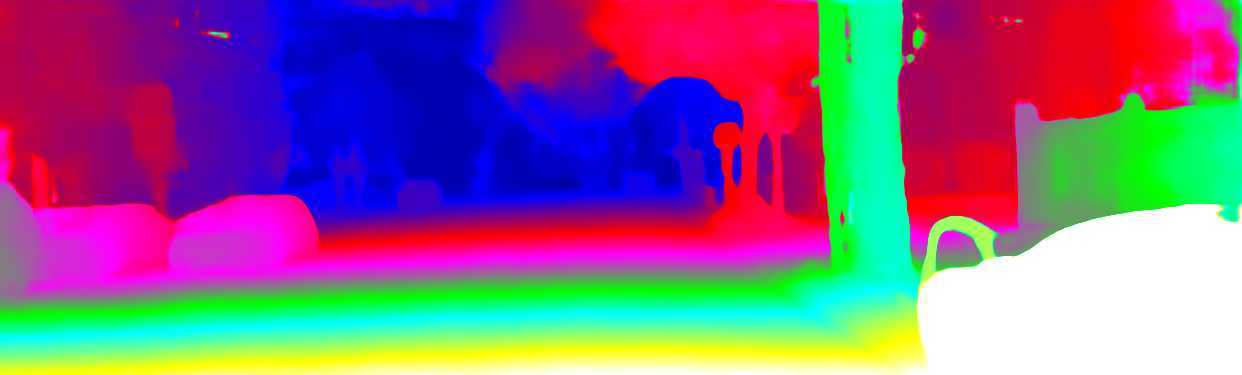}
	\includegraphics[width=0.31\linewidth]{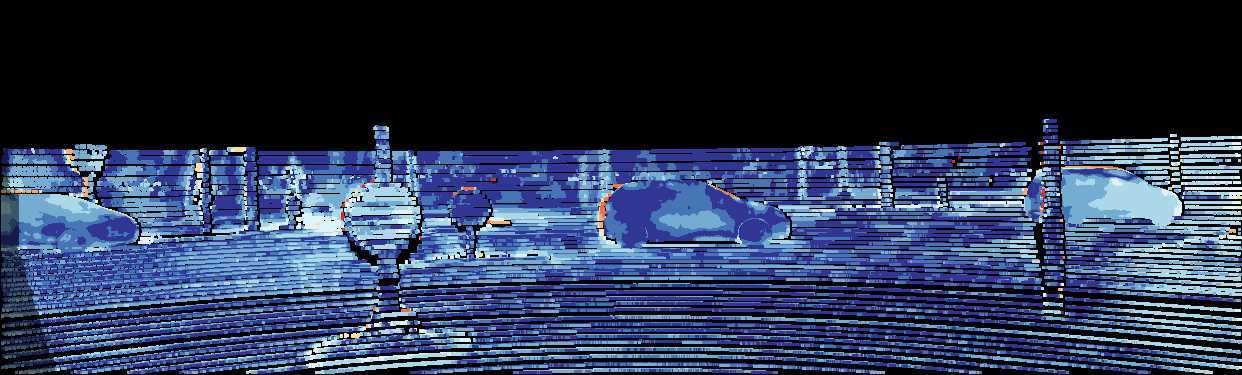}
	\includegraphics[width=0.31\linewidth]{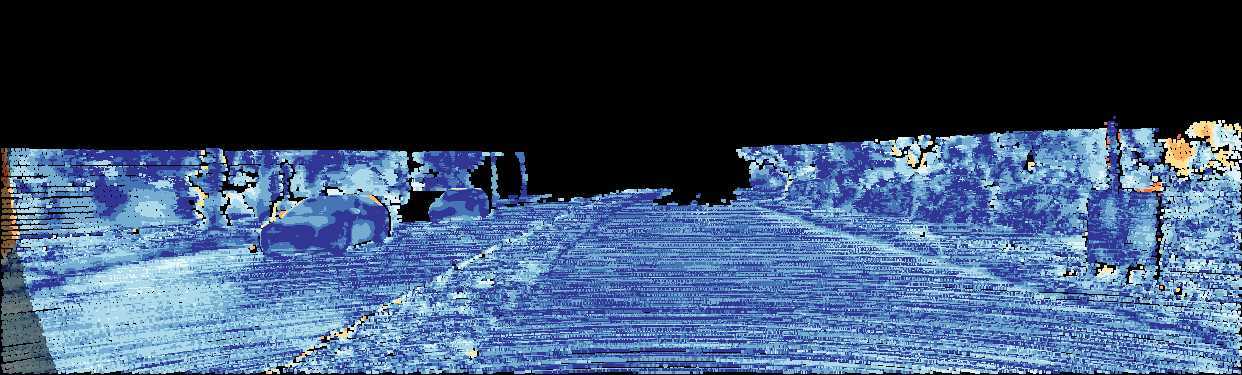}
	\includegraphics[width=0.31\linewidth]{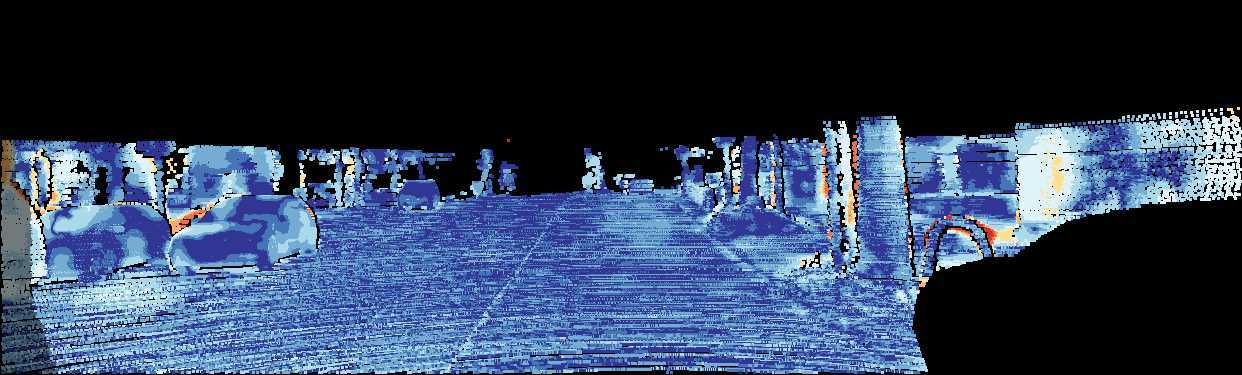}
	\includegraphics[width=0.31\linewidth]{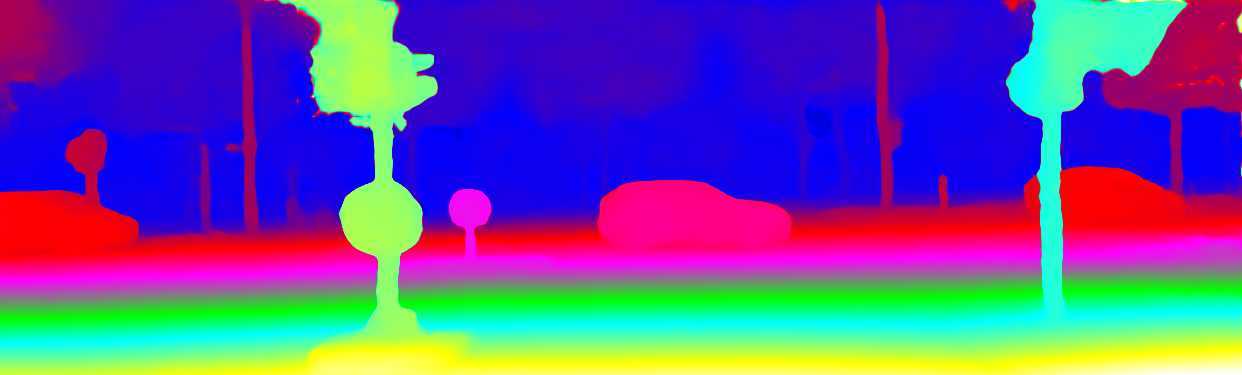}
	\includegraphics[width=0.31\linewidth]{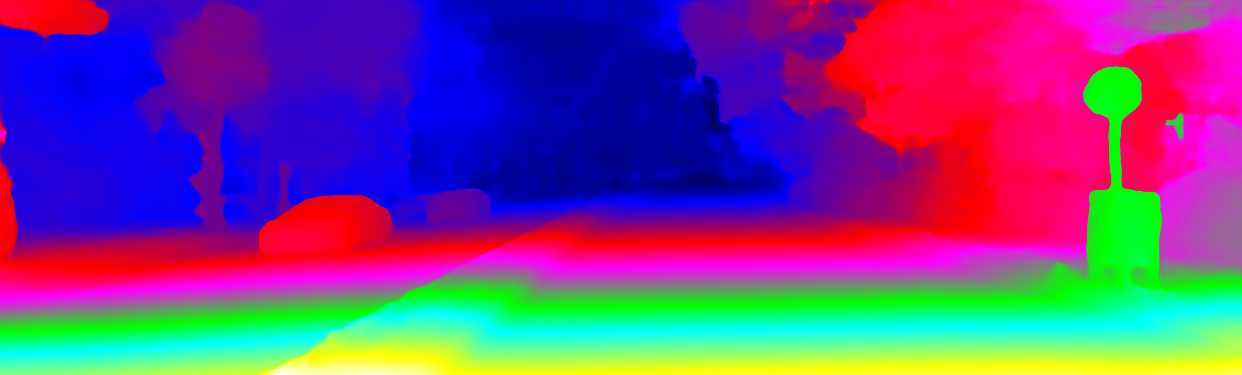}
	\includegraphics[width=0.31\linewidth]{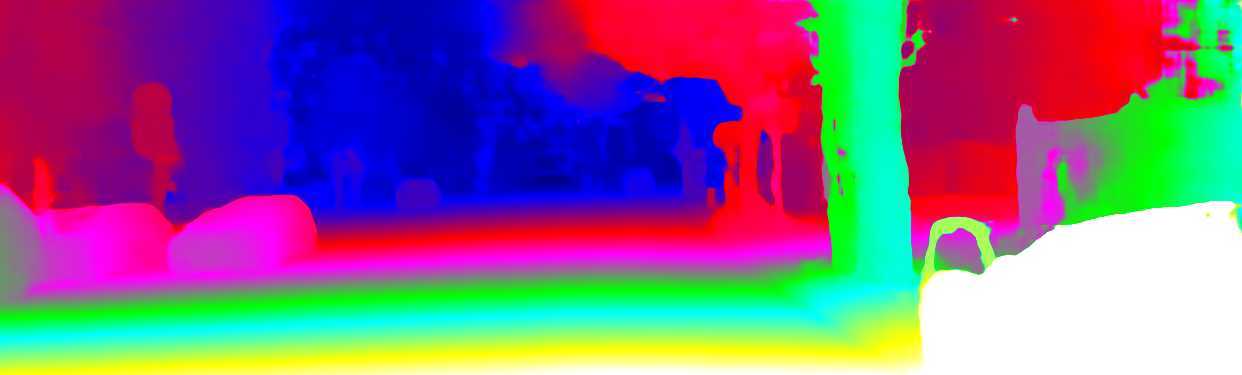}
	\includegraphics[width=0.31\linewidth]{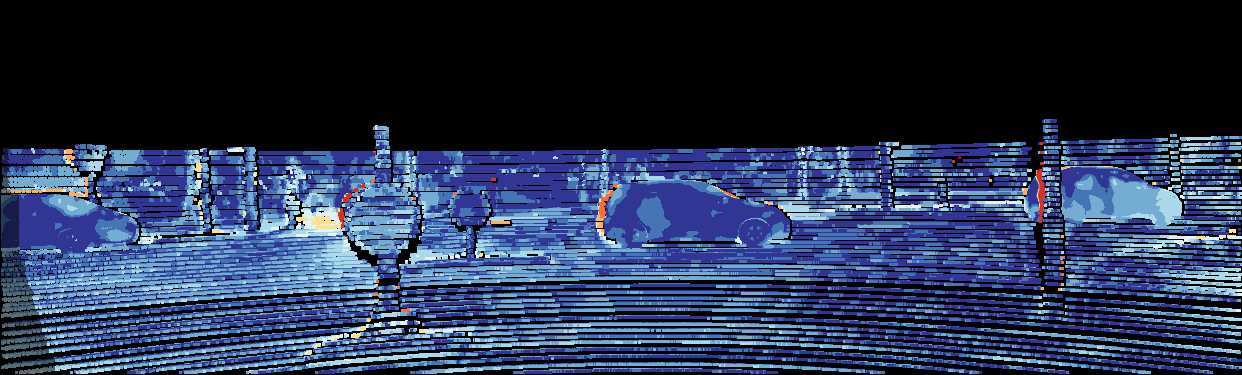}
	\includegraphics[width=0.31\linewidth]{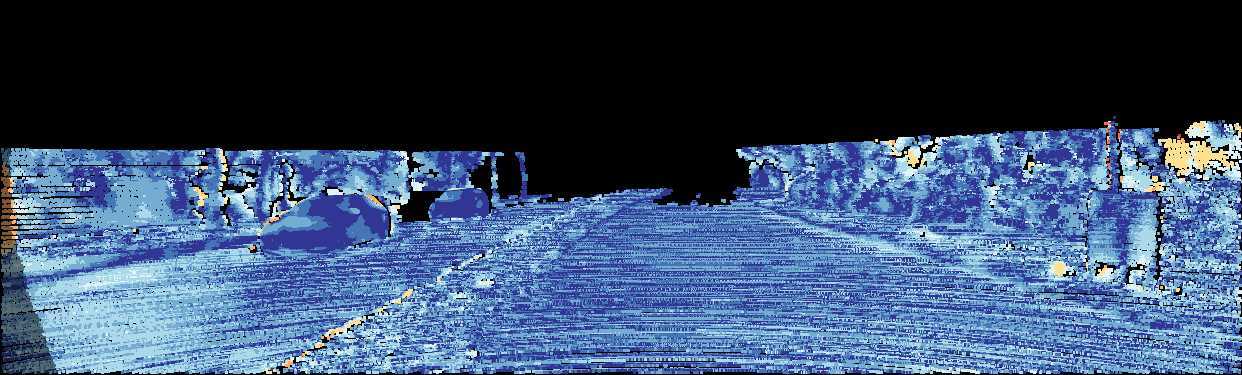}
	\includegraphics[width=0.31\linewidth]{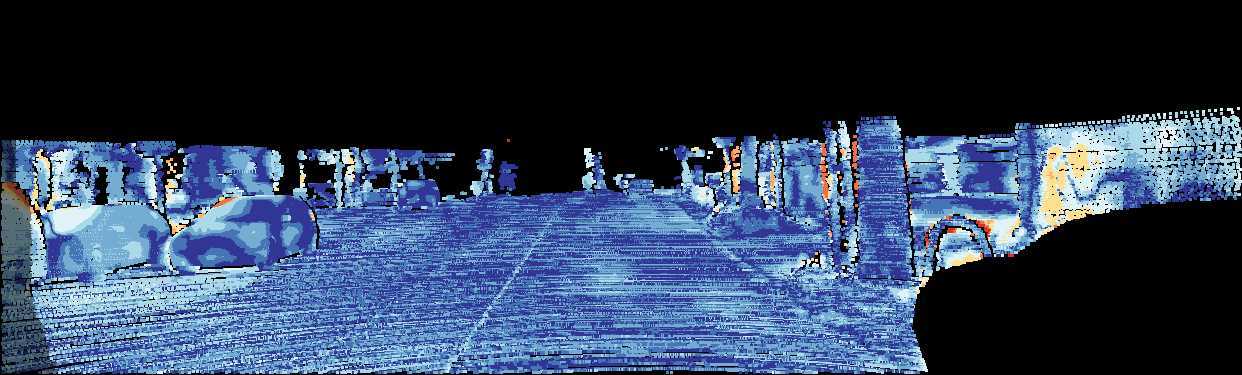}
	\caption{\textbf{KITTI 2015 test data qualitative results.} From top: left stereo input image, our disparity prediction, our error map, PSMNet's disparity predicetion, its error map}
	\label{fig:Kitti2015}
\end{figure*}

\subsection{Disparity Resgression}
Traditionally, the stereo algorithm can estimate the depth offset of the cost volume by argmin operation, which failed to produce sub-pixel disparity estimatation and be trained using back-propagation. In order to obtain more robust and effective disparity map by regression, we exploit soft argmin to accomplish the disparity map regression\cite{Kendall2017}. Above all, we calculate the probability of each disparity $d$ converted from the cost volume \(c\) via the softmax operation \(\sigma\left( \cdot  \right) \). Second, we take the sum of each disparity \(\hat{d} \), weighted by its normalized probability. The corresponding mathematical formula is as (1):
\begin{eqnarray}
\hat{d} = \sum\limits_{d = 0}^{{D_{\max }}} {d \times \sigma \left(  - c_d  \right)}
\end{eqnarray}

\subsection{Loss Function}
As many recent stereo matching methods mentioned, smooth  \(L_1\) loss is widely used in bounding box regression for object detection beacuse of its robustness and low sensitivity to outliers. So we set the loss function of CFP-Net as
\begin{eqnarray}
L\left (d, \hat{d} \right )=\frac{1}{N} \sum_{i=1}^{N}\mathit{Smooth_{L_1}}\left (d_i - \hat{d_i}\right )
\end{eqnarray}
in which
\begin{eqnarray}
\mathit{Smooth_{L_1}\left( x \right) } =
\left\{
\begin{array}{lll}
\frac{1}{2}x^2, \ \ if \left| x \right| < 1. \\
 \left| x \right|-0.5, \ \ otherwise.
\end{array}
\right.
\end{eqnarray}

where $N$ is the number of all labeled pixels, \(\hat{d_i}\) is the predicted disparity and $d_i$ is the ground-truth disparity.

\section{Experiments}
We will present out experimental details and results in this section. In section 4.1 we show our experimental datasets and training parameters. We show the performance of proposed CFP-Net on recognized stereo datasets from different perspectives. In addition to this, We conduct multiple comparative experiments for comparing our proposed CFSPP with ASPP and SPP in section 4.2. Then we compare the proposed method with other state-of-the-art published methods. Finally, we also visualize the disparity maps generated by CFP-Net and compare with others on KITTI stereo 2012\cite{6248074}, KITTI stereo 2015\cite{Menze2015CVPR} and Scene Flow\cite{MIFDB16} in section 4.3.

\subsection{Dataset descrption}
Following are the main recognized stereo datasets used during training.

(i) KITTI: KITTI datasets are real-world datasets with street views from a driving car, containing KITTI-2012 and KITTI-2015 stereo datasets. Both stereo image pair and ground-truth disparity have a resolution with \(376 \times 1280\). 

(ii) Scene Flow: Scene Flow is a synthetic dataset, containing 35454 training stereo image pairs and 4370 testing stereo image pairs, with all image pair having dense and elaborate ground-truth disparities. Both training set and validation set have a resolution of \(540 \times 960\).

\begin{figure*}[!htb]
	\centering
	\includegraphics[width=0.31\linewidth]{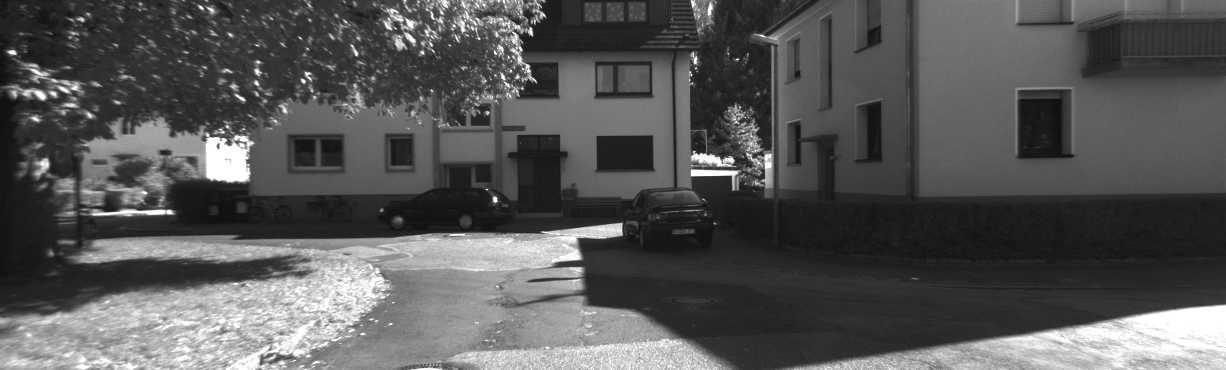}
	\includegraphics[width=0.31\linewidth]{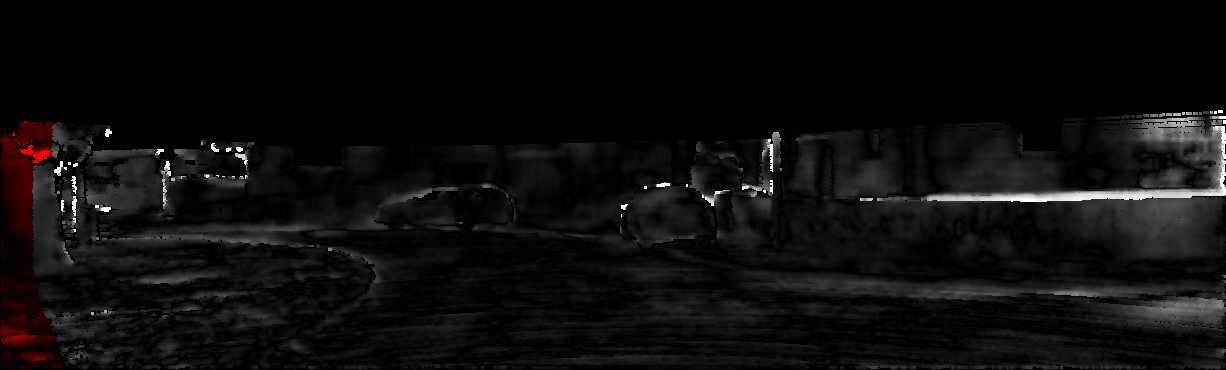}
	\includegraphics[width=0.31\linewidth]{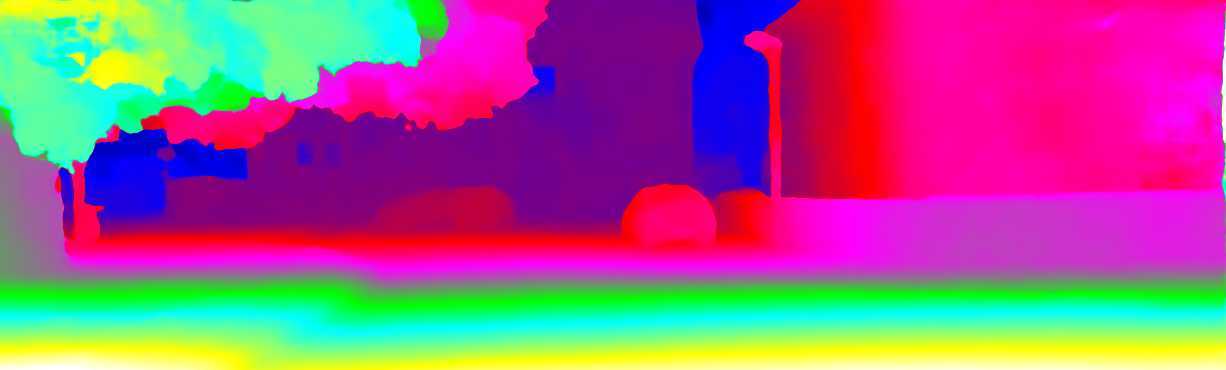}
	\includegraphics[width=0.31\linewidth]{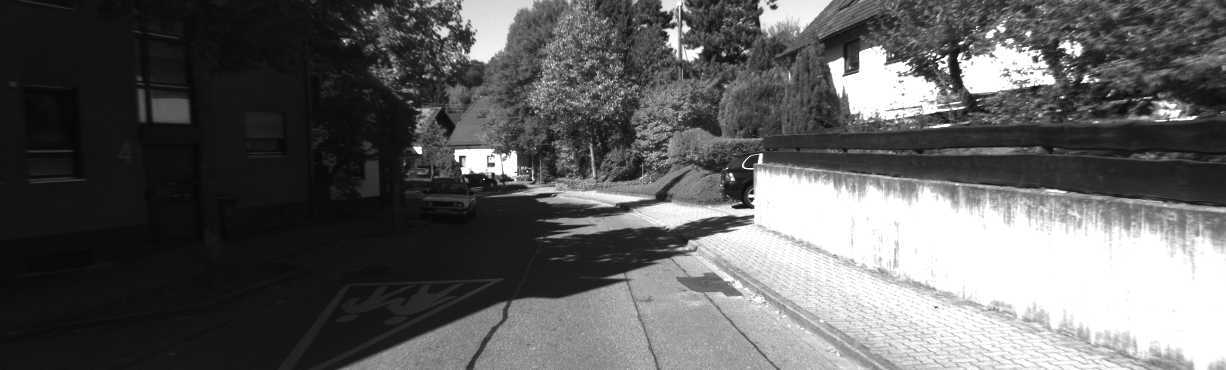}
	\includegraphics[width=0.31\linewidth]{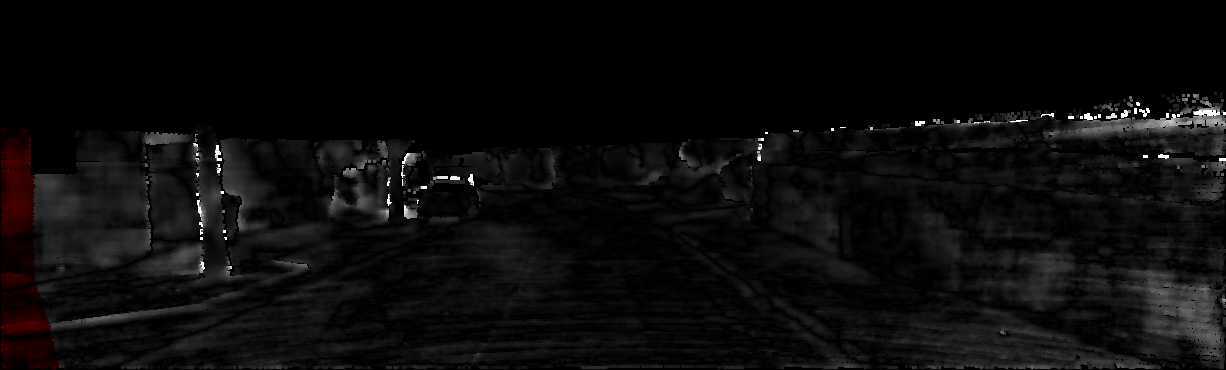}
	\includegraphics[width=0.31\linewidth]{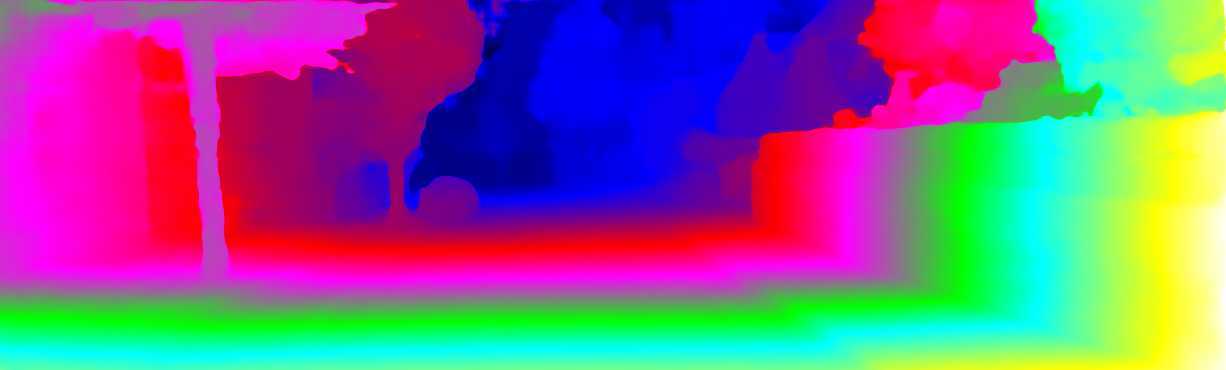}
	\caption{\textbf{KITTI 2012 test data qualitative results.} From left: left stereo input image, disparity prediction, error map.}
	\label{fig:Kitti2012}
\end{figure*}

\subsection{Comparative Experiments}
\label{sec:Comparative Experiments}
In this section, we will conduct several comparative experiments with different settings to evaluate the effectiveness of the specific structure in CFP-Net on Scene Flow dataset. 

\textbf{\small Comparative Experiments for Context Pyramid Module.} As described in section 3.3, we conduct multiple comparative experiments to compare SPP, ASPP and proposed CFSPP mentioned above. Specific experimental results are shown in Table \ref{tab:Pivotal Architecture Comparison}. The result shows that the CFSPP proposed performs better than other two structure on Scene Flow datasets. It proves that CFSPP can gather more abundant and detailed context information comparing SPP and ASPP. We think CFSPP can combine multi-level context information of the object on the same scale from different perspectives.

\textbf{\small Comparative Experiments for Local Feature Extraction and 3DCNN Module.}
For evaluating the effectiveness of the local feature extraction module and 3DCNN module, we replace all the layers of LFE and 3DCNN with conventional 2D or 3D convolutional layers. Specific experimental results are shown in Tab. \ref{tab:Pivotal Architecture Comparison}. Some typical samples of SceneFlow datasets are shown in Fig. \ref{fig:SceneFlow}.

\subsection{Results on KITTI}
We applied our best model to calculate the final disparity maps for the testing images in the KITTI dataset, then submitted our results to KITTI online evaluation server. Recent methods' evaluation results(only for published papers, including ours) on KITTI leaderboard are shown in Tab. \ref{tab:Model Performance on 2015} and Tab. \ref{tab:Model Performance on 2012}. As shown on Tab. \ref{tab:Model Performance on 2015}, we exploited the percentage of erroneous pixels in the background (D1-bg), foreground (D1-fg) and all pixels (D1-all) in the two areas mentioned above to evaluate our model's performance quantitatively. For explicitly demonstrating our method's effectiveness and practicality better, as shown in Fig. \ref{fig:Kitti2015}, we picked out some typical samples of the disparity maps predicted by the PSMNet, together with corresponding our error maps from KITTI online evaluation server. Obviously, the proposed CFP-Net performs better than others in textureless and foreground regions on KITTI 2015 datasets.

Tab. \ref{tab:Model Performance on 2012} illustrates our's together with recent state-of-the-art methods' evaluation results on KITTI 2012 datasets. We perform well both in non-occluded (Out-Noc) and all (Out-All) area. Moreover, our method is also highly efficient because of our succinct network with only 4.6M parameters and program optimization. It is tested that the overall run time is only 0.95 seconds on a single Nvidia Titan X GPU.

\section{Conclusions}
In this work, we propose CFP-Net, a novel network architecture consists of the local feature extraction module(LFE), cross-form spatial pyramid module(CFSPP) and multi-scale 3D convolution matching and fusion module. It performs well in ill-posed regions for its' enlarged receptive fields and better utilization of context information in multi-scale cost volume. However, recent stereo matching methods based CNN still perform poorly on estimating disparity for inherently ill-posed and textureless regions. It is still an intractable problem that how to incorporate different scales and locations of feature maps better and regularize cost volume better. In our future work, we will ameliorate our cost volume reconstruction and refinement network. 

\section*{Acknowledgment}
This work was supported in part by Natural Science Foundation of China (61420106007, 61671387)

\bibliography{Cite}

\begin{thebibliography}{10}

\bibitem{Chang2018Pyramid}
Jiaren Chang and Yongsheng Chen.
\newblock {Pyramid Stereo Matching Network}.
\newblock In {\em arXiv preprint}, 2018.

\bibitem{Chen2018Encoder}
Liangchieh Chen, Yukun Zhu, George Papandreou, Florian Schroff, and Hartwig
  Adam.
\newblock Encoder-decoder with atrous separable convolution for semantic image
  segmentation.
\newblock 2018.

\bibitem{6248074}
A.~Geiger, P.~Lenz, and R.~Urtasun.
\newblock Are we ready for autonomous driving? the kitti vision benchmark
  suite.
\newblock In {\em 2012 IEEE Conference on Computer Vision and Pattern
  Recognition}, pages 3354--3361, June 2012.

\bibitem{Guney2015}
Fatma Guney and Andreas Geiger.
\newblock {Displets: Resolving stereo ambiguities using object knowledge}.
\newblock In {\em Proceedings of the IEEE Computer Society Conference on
  Computer Vision and Pattern Recognition}, 2015.

\bibitem{He2015}
Kaiming He, Xiangyu Zhang, Shaoqing Ren, and Jian Sun.
\newblock {Spatial Pyramid Pooling in Deep Convolutional Networks for Visual
  Recognition}.
\newblock {\em IEEE Transactions on Pattern Analysis and Machine Intelligence},
  2015.

\bibitem{4359315}
H.~Hirschmuller.
\newblock Stereo processing by semiglobal matching and mutual information.
\newblock {\em IEEE Transactions on Pattern Analysis and Machine Intelligence},
  30(2):328--341, Feb 2008.

\bibitem{Kendall2017}
Alex Kendall, Hayk Martirosyan, Saumitro Dasgupta, and Peter Henry.
\newblock {End-to-End Learning of Geometry and Context for Deep Stereo
  Regression}.
\newblock In {\em 2017 IEEE International Conference on Computer Vision
  (ICCV)}, volume 2017-Octob, pages 66--75. IEEE, oct 2017.

\bibitem{Klaus2006Segment}
Andreas Klaus, Mario Sormann, and Konrad Karner.
\newblock Segment-based stereo matching using belief propagation and a
  self-adapting dissimilarity measure.
\newblock In {\em Proc. IEEE International Conference on Pattern Recognition,
  Sep}, pages 15--18, 2006.

\bibitem{Kolmogorov2001Computing}
Vladimir Kolmogorov and Ramin Zabih.
\newblock {\em Computing Visual Correspondence with Occlusions using Graph
  Cuts}.
\newblock Cornell University, 2001.

\bibitem{LI2018328}
Bo~Li, Yuchao Dai, and Mingyi He.
\newblock Monocular depth estimation with hierarchical fusion of dilated cnns
  and soft-weighted-sum inference.
\newblock {\em Pattern Recognition}, 83:328--339, 2018.

\bibitem{Li2015Depth}
Bo~Li, Chunhua Shen, Yuchao Dai, Anton Van~Den Hengel, and Mingyi He.
\newblock Depth and surface normal estimation from monocular images using
  regression on deep features and hierarchical crfs.
\newblock In {\em Computer Vision and Pattern Recognition}, pages 1119--1127,
  2015.

\bibitem{MIFDB16}
N.~Mayer, E.~Ilg, P.~H{\"a}usser, P.~Fischer, D.~Cremers, A.~Dosovitskiy, and
  T.~Brox.
\newblock A large dataset to train convolutional networks for disparity,
  optical flow, and scene flow estimation.
\newblock In {\em IEEE International Conference on Computer Vision and Pattern
  Recognition (CVPR)}, 2016.
\newblock arXiv:1512.02134.

\bibitem{Menze2015CVPR}
Moritz Menze and Andreas Geiger.
\newblock Object scene flow for autonomous vehicles.
\newblock In {\em Conference on Computer Vision and Pattern Recognition
  (CVPR)}, 2015.

\bibitem{Pang2017Cascade}
Jiahao Pang, Wenxiu Sun, Jimmy~Sj Ren, Chengxi Yang, and Qiong Yan.
\newblock Cascade residual learning: A two-stage convolutional neural network
  for stereo matching.
\newblock pages 878--886, 2017.

\bibitem{Seki2016Patch}
Akihito Seki and Marc Pollefeys.
\newblock Patch based confidence prediction for dense disparity map.
\newblock In {\em British Machine Vision Conference}, pages 23.1--23.13, 2016.

\bibitem{Seki2017SGM}
Akihito Seki and Marc Pollefeys.
\newblock Sgm-nets: Semi-global matching with neural networks.
\newblock In {\em IEEE Conference on Computer Vision and Pattern Recognition},
  pages 6640--6649, 2017.

\bibitem{Shaked2017Improved}
Amit Shaked and Lior Wolf.
\newblock Improved stereo matching with constant highway networks and
  reflective confidence learning.
\newblock In {\em IEEE Conference on Computer Vision and Pattern Recognition},
  pages 6901--6910, 2017.

\bibitem{Zhang2018Multi}
Jing Zhang, Yuchao Dai, Fatih Porikli, and Mingyi He.
\newblock Multi-scale salient object detection with pyramid spatial pooling.
\newblock In {\em Asia-Pacific Signal and Information Processing Association
  Summit and Conference}, pages 1286--1291, 2018.

\bibitem{Zhong2017}
Yiran Zhong, Yuchao Dai, and Hongdong Li.
\newblock {Self-Supervised Learning for Stereo Matching with Self-Improving
  Ability}.
\newblock In {\em arXiv preprint}, 2017.

\end{thebibliography}

\end{document}